\documentclass[runningheads]{llncs}
\usepackage{graphicx}
\usepackage{amsmath}
\usepackage{amsfonts}
\usepackage{multirow}
\usepackage{booktabs}
\usepackage{xcolor}
\usepackage[export]{adjustbox}
\usepackage{algorithm}
\usepackage{algpseudocode}
\usepackage{tabularx}
\usepackage{ bbold }
\usepackage{hyperref}
\newcommand{\etal}{\textit{et al.}}
\newcommand{\ie}{\textit{i.e.}}

\newcommand{\mysubtitle}[1]{\noindent{\bf #1}}

\begin{document}
%
\title{Learning to Rank Words:\\ Optimizing Ranking Metrics for Word Spotting}
%
%
\author{Pau Riba\orcidID{0000-0002-4710-0864} \and Adrià Molina\orcidID{0000-0003-0167-8756} \and Lluis Gomez\orcidID{0000-0003-1408-9803} \and Oriol Ramos-Terrades\orcidID{0000-0002-3333-8812} \and Josep Lladós\orcidID{0000-0002-4533-4739}}
%
\authorrunning{P. Riba et al.}
%
\institute{Computer Vision Center and Computer Science Department, \\ Universitat Autònoma de Barcelona, Catalunya\\ \email{\{priba,lgomez,oriolrt,josep\}@cvc.uab.cat},\\ adria.molinar@e-campus.uab.cat}
%
\maketitle              
\begin{abstract}
    In this paper, we explore and evaluate the use of ranking-based objective functions for learning simultaneously a word string and a word image encoder. We consider retrieval frameworks in which the user expects a retrieval list ranked according to a defined relevance score. In the context of a word spotting problem, the relevance score has been set according to the string edit distance from the query string. We experimentally demonstrate the competitive performance of the proposed model on query-by-string word spotting for both, handwritten and real scene word images. We also provide the results for query-by-example word spotting, although it is not the main focus of this work.

\end{abstract}

\section{Introduction}
Word spotting, also known as keyword spotting, was introduced in the late 90's in the seminal papers of Manmatha~\etal~\cite{manmatha1996wordspotting,manmatha1996indexing}. It emerged quickly as a highly effective alternative to text recognition techniques in those scenarios with scarce data availability or huge style variability, where a strategy based on full transcription is still far from being feasible and its objective is to obtain a ranked list of word images that are relevant to a user's query. Word spotting has been typically classified in two particular settings according to the target database gallery. On the one hand, there are the segmentation-based methods, where text images are segmented at word image level~\cite{gomez2017lsde,sudholt2016phocnet}; and, on the other hand, the segmentation-free methods, where words must be spotted from cropped text-lines, or full documents~\cite{almazan2014segmentation,rusinol2015efficient}. Moreover, according to the query modality, these methods can be classified either  query-by-example (QbE)~\cite{rusinol2011browsing} or query-by-string (QbS)~\cite{almazan2014word,krishnan2016deep,sudholt2016phocnet,wilkinson2016semantic}, being the second one, the more appealing from the user perspective. 


The current trend in word spotting methods is based on learning a mapping function from the word images to a known word embedding spaces that can be handcrafted~\cite{sudholt2016phocnet,wilkinson2016semantic} or learned by another network~\cite{gomez2017lsde}. These family of approaches have demonstrated a very good performance in the QbS task. However, this methods focus on optimizing this mapping rather than making use of a ranking-based learning objective which is the final goal of a word spotting system. 

In computer vision, siamese or triplet networks~\cite{weinberger2009distance} trained with margin loss have been widely used for the image retrieval task. Despite their success in a large variety of tasks, instead of considering a real retrieval, these architectures act locally in pairs or triplets. To overcome this drawback, retrieval-based loss functions have emerged. These novel objective functions aim to optimize the obtained ranking. For instance, some works have been proposed to optimize the average precision (AP) metric~\cite{brown2020eccv,revaud2019learning}. Other works, such as Li~\etal~\cite{Li2021PNP} suggested to only consider negative instances before positive ones to learn ranking-based models. Traditionally, word spotting systems are evaluated with the mean average precision (mAP) metric. However, this metric only takes into account a binary relevance function. Besides mAP, in these cases where a graded relevance score is required, information retrieval research has proposed the normalized Discounted Cumulative Gain (nDCG) metric. This is specially interesting for image retrieval and, in particular, word spotting because from the user perspective, the expected ranking list should follow such an order that is relevant to the user even though not being a perfect match. This metric has also been previously explored as a learning objective~\cite{valizadegan2009learning}.

\begin{figure}
    \centering
    \includegraphics[width=\textwidth]{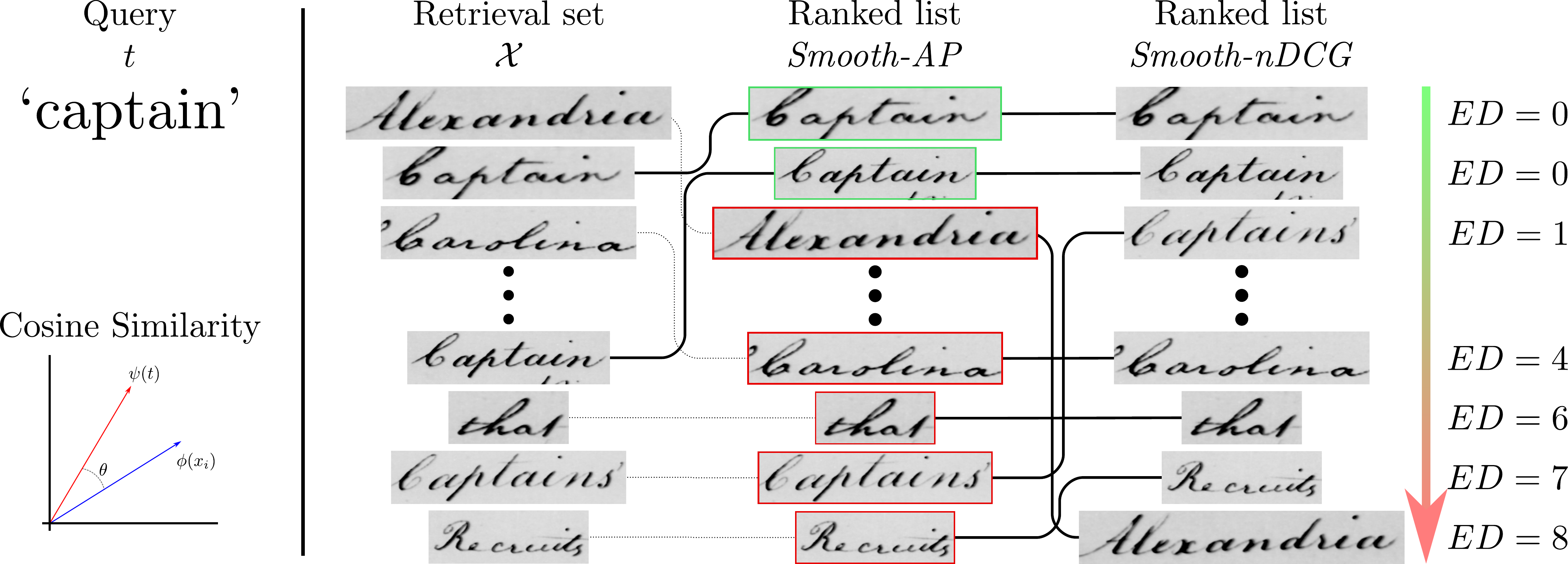}
    \caption{Overview of the behavior of the proposed model. Given a query \(t\) and the retrieval set $\mathcal{X}$, the proposed word spotting system uses the cosine similarity to rank the retrieval set accordingly. Observe that Smooth-AP considers a binary relevance score whereas in the case of Smooth-nDCG, the ranked list is graded according a non-binary relevance score such as the string edit distance.}
    \label{f:summary}
\end{figure}

Taking into account the previous comments, the main motivation of this work is to analyze and evaluate two retrieval-based loss functions, namely Smooth-AP~\cite{brown2020eccv} and Smooth-nDCG, to train our QbS word spotting system based on the mAP and nDCG metrics respectively. We hypothesize that, with different purposes from the application perspective, training a word spotting system according to its retrieval list should lead to more robust embeddings. Figure~\ref{f:summary} provides an overview of the expected behavior for each loss. On the one hand, the Smooth-AP loss expects to find the relevant images at the top ranking position~\ie~these words with the same transcription as the query. On the other hand, Smooth-nDCG aims at obtaining a more interpretable list from the user's perspective. For instance, the graded relevance function can consider the string edit distance or any other semantic similarity among strings depending on the final application.

To summarize, the main contributions of this work are:
\begin{itemize}
    \item We propose a word spotting system which is trained solely with retrieval-based objective functions.
    \item We introduce a novel ranking loss function, namely, \emph{Smooth-nDCG}, which is able to train our system according to a known relevance feedback. Therefore, we can present the results in a more pleasant way.
    \item We conduct extensive evaluation to demonstrate, on the one hand, the advantages and disadvantages of the proposed learning objectives and, on the other hand, the boost in word spotting performance for the QbS settings.
\end{itemize}

The rest of this paper is organized as follows. Section~\ref{s:related} reviews the previous works that are relevant to the current task. Afterwards, Section~\ref{s:architecture} presents the embedding models for both, images and strings. Section~\ref{s:experiments} conducts an extensive evaluation on the proposed word spotting system. Finally, Section~\ref{s:conclusions} draws the conclusions and the future work.

\section{Related Work}
\label{s:related}

\subsection{Word Spotting}

In this section, we introduce word spotting approaches that are relevant to the current work.  Word spotting has been a hot research topic from its origin. The first successful attempts on using neural networks for word spotting were done by Frinken~\etal~\cite{frinken2011novel}. They adapted the CTC Token passing algorithm to be applied to word spotting. In that work, the authors apply a BLSTM network to segmented text line images and then the CTC Token passing algorithm. It's main limitation is that it can only perform QbS tasks but not QbE.

Later, the research challenge evolved to integrate visual and textual information to perform the retrieval in a QbS setting. In this regard, one of the first works proposing this integration was Aldavert~\etal~\cite{aldavert2013integrating}. In their work, the textual description is built based on n-gram histograms while the visual description is built by a pyramidal representation of bag of words~\cite{Lazebnik2006Pyramidal}. Both representations are then projected to a common space by means of Latent Semantic Analysis (LSA)~\cite{deerwester1990LSA}.


Another relevant work, introduced by Almazan~\etal~\cite{almazan2014word}, propose a pyramidal histogram of characters, which they called PHOC, to represent in the same embedding space segmented word images and their corresponding text transcription. This shared representation enables exchange the input and output modalities in QbS and QbE scenarios. This seminal work inspired the current state-of-the-art methods~\cite{sudholt2016phocnet,wilkinson2016semantic,krishnan2016deep} on using the PHOC representation. In~\cite{sudholt2016phocnet}, Sudholt~\etal propose the PHOCNet architecture to learn an image embedding representation of segmented words. This network architecture adapts the previous PHOC representation to be learned using a deep architecture. To deal with changing image sizes, the authors make us of a spatial pyramid pooling layer~\cite{he2015spatial}. Similarly, Wilkinson~\etal~\cite{wilkinson2016semantic} train a triplet CNN followed a fully connected 2-layer network to learn an image embedding representation in the space of word embeddings. They evaluate two hand-crafted embeddings, the PHOC and the discrete cosine transform. To learn a shared embedding space, they used the cosine loss to train the network.



Krishnan~\etal~\cite{krishnan2016deep} also proposed a deep neuronal architecture to learn a common embedding space for visual and textual information. The authors use a pre-trained CNN on synthetic handwritten data for segmented word images~\cite{Krishna2016eccv} and the PHOC as attribute-based text representation. Later, they presented an improved version of this work in \cite{krishnan2018word}. On the one hand, the architecture is improved to the ResNet-34. they also exploit the use of synthetic word images at test time.

However, all these methods focus on training architectures to rank first relevant images but without ranking them into a  retrieval list.

More recently, Gomez~\etal~\cite{gomez2017lsde} faced the problem of obtaining a more appealing retrieval list. First, they proposed a siamese network architecture that is able to learn a string embedding that correlates with the Levenshtein edit distance~\cite{levenshtein1966binary}. Second, they train a CNN model inspired from~\cite{Jaderberg2014} to train an image embedding close to the corresponding string embedding. Thus, even though they share our motivation, they do not exploit this ranking during training.

\subsection{Ranking metrics}

In word spotting and, more generally in information retrieval, several metrics have been carefully designed to evaluate the obtained rankings. We briefly define the two main ones, that are considered in this work.

\mysubtitle{Mean Average Precision}: Word spotting performance has been traditionally measured by the \emph{mean Average Precision} (mAP), a classic information retrieval metric~\cite{rusinol2009performance}. The {\em Average Precision} given a query $q$ ($\operatorname{AP}_q$) is formally defined as the mean

\begin{equation}\label{eq:ap}
    \operatorname{AP}_q = \frac{1}{|\mathcal{P}_q|} \sum_{n=1}^{|\Omega_q|}P@n \times r(n),
\end{equation}

\noindent where \(P@n\) is the precision at \(n\), \(r(n)\) is a binary function on the relevance of the \textit{n}-th item in the returned ranked list, $\mathcal{P}_q$ 
is the  set of all relevant 
objects with regard the query $q$ and \(\Omega_q
\) is the set of retrieved elements from the dataset. Then, the mAP is defined as:
\begin{equation}
    \operatorname{mAP} = \frac{1}{Q}\sum_{q=1}^{Q} \operatorname{AP}_q,
\end{equation}
where $Q$ is the number of queries.

\mysubtitle{Normalized Discounted Cumulative Gain}: In information retrieval, the \emph{normalized Discounted Cumulative Gain} (nDCG) is used to measure the performance on such scenarios where instead of a binary relevance function, we have a graded relevance scale. The main idea is that highly relevant elements appearing lower in the retrieval list should be penalized. The \emph{Discounted Cumulative Gain} (DCG) for a query \(q\) is defined as %
\begin{equation}\label{eq:dcg}
    \operatorname{DCG}_q = \sum_{n=1}^{|\Omega_q|} \frac{r(n)}{\log_2(n+1)},
\end{equation}%
where \(r(n)\) is a graded function on the relevance of the \textit{n}-th item in the returned ranked list and $\Omega_q$ is the set of retrieved elements as defined above. However, to allow a fair comparison among different queries, a normalized version was proposed and defined as
\begin{equation}\label{eq:ndcg}
    \operatorname{nDCG}_q = \frac{\operatorname{DCG}_q}{\operatorname{IDCG}_q},
\end{equation}
where $\operatorname{IDCG}_q$ is the ideal discounted cumulative gain, \ie~assuming a perfect ranking according to the relevance function. It is formally defined as
\begin{equation}
    \operatorname{IDCG}_q = \sum_{n=1}^{|\Lambda_q|} \frac{r(n)}{\log_2(n+1)}
    \label{eq:idcg}
\end{equation} %
where \(\Lambda_q\) is the ordered set according the relevance function.

\section{Architecture}
\label{s:architecture}

\subsection{Problem formulation}

Let \(\{\mathcal{X}, \mathcal{Y}\}\) be a word image dataset, containing word images \(\mathcal{X}=\{x_i\}_{i=0}^N\), and their corresponding transcription strings \(\mathcal{Y}\). Let \(\mathcal{A}\) be the alphabet containing the allowed characters. Given the word images \(\mathcal{X}\) as a retrieval set or gallery for searching purposes on the one hand, and a given text string \(t\) such that its characters \(t_i\in\mathcal{A}\) on the other hand; the proposed embedding functions \(\phi(\cdot)\) and \(\psi(\cdot)\) for word images and text strings respectively, have the objective to map its input in such a space that a given similarity function \(S(\cdot,\cdot)\) is able to retrieve a ranked list with the relevant elements. Traditionally, the evaluation of word spotting divides this list in two partitions, namely the positive or relevant word images \ie~their transcription matches with the query \(t\), and the negative or non-relevant word images. Thus, the aim of a word spotting system is to rank the positive elements at the beginning of the retrieval list. However, from the user perspective, the non-relevant elements might be informative enough to require them to follow some particular order. Therefore, we formulate the word spotting problem in terms of a relevance score \(R(\cdot,\cdot)\) for each element in the list. Finally, given a query \(t\) we can formally define this objective as
\begin{equation}
S(\phi(x_i), \psi(t)) > S(\phi(x_j), \psi(t)) \Longleftrightarrow R(y_i, t) > R(y_j, t),
\end{equation}
where \(x_i, x_j \in \mathcal{X}\) are two elements in the retrieved list and \(y_i, y_j \in \mathcal{Y}\) their corresponding transcriptions.

\subsection{Encoder Networks}

As already explained, the proposed word spotting system consists of a word image encoding \(\phi(\cdot)\) and a textual encoding \(\psi(\cdot)\). The outputs of these encoding functions are expected to be in the same space.

\mysubtitle{Word image encoding}: Given a word image \(x_i\in \mathcal{X}\), it is firstly resized into a fixed height, but we keep the original aspect ratio. As the backbone of our image encoding, we use a ImageNet pre-trained ResNet-34~\cite{he2016deep} network. As a result of the average pooling layer, our model is able to process images of different widths. Finally, we \(L_2\)-normalize the final 64D embedding. 

\mysubtitle{String encoder}: The string encoder embeds a given query string \(t\) in a 64D space.  Firstly, we define a character-wise embedding function \(e \colon \mathcal{A} \to \mathbb{R}^m\). Thus, the first step of our encoder is to embed each character \(c \in t\) into the character latent space. Afterwards, each character is feed into a Bidirectional Gated Recurrent Unit (GRU)~\cite{chung2014empirical} with two layers. In addition, a linear layer takes the last hidden state of each direction to generate our final word embedding. Finally, we also \(L_2\)-normalize the final 64D embedding. 

Although both embeddings can be compared by means of any arbitrary similarity measure \(S(\cdot, \cdot)\), in this work we decided to use the cosine similarity between two embeddings \(v_q\) and \(v_i\), which is defined as:
\begin{equation}
S(v_q, v_i) = \frac{v_q \cdot v_i}{\|v_q\|\|v_i\|}.
\end{equation}

From now on and for the sake of simplicity, given a query \(q\) its corresponding similarity against the \(i\)-th element of the retrieval set is denoted as \(s_i = S(v_q, v_i)\).

\subsection{Learning Objectives}

As already stated in the introduction, we analyze two learning objectives which provide supervision at the retrieval list level rather than at sample, pair or triplet level. The proposed losses are inspired by the classical retrieval evaluation metrics introduced in Section~\ref{s:related}.

Let us first define some notations and concepts.

\mysubtitle{Ranking Function} ($\mathcal{R}$). Following the notation introduced in~\cite{qin2010general}, these information retrieval metrics can be reformulated by means of the following ranking function,
\begin{equation}
    \mathcal{R}(i, \mathcal{C}) = 1 + \sum_{j=1}^{|\mathcal{C}|} \mathbb{1}\{D_{ij}<0\},
    \label{eq:ranking}
\end{equation}
where $\mathcal{C}$ is any set (such as $\Omega_q$ or $\mathcal{P}_q$), $D_{ij}=s_i-s_j$ and \(\mathbb{1}\{\cdot\}\) is an Indicator function.

However, with this formulation, we are not able to optimize following the gradient based optimization methods and an smooth version of the Indicator is required. Even though several approximations exist, in this work we followed the one proposed by Quin~\etal~\cite{qin2010general}, which  make use of the sigmoid function 
\begin{equation}
    \mathcal{G}(x;\tau) = \frac{1}{1+e^\frac{-x}{\tau}}.
    \label{eq:sigmoid}
\end{equation}

\mysubtitle{Smooth-AP}: The smoothed approximation of AP, namely Smooth-AP and proposed by Brown~\etal~\cite{brown2020eccv}, has shown a huge success on image retrieval. There, the authors replace the $P@n \times r(n)$ term in Equation \ref{eq:ap} by the ranking function and the exact AP equation becomes 
\begin{equation}
    \operatorname{AP}_q = \frac{1}{|\mathcal{P}_q|} \sum_{i\in \mathcal{P}_q} \frac{1 + \sum_{j\in \mathcal{P}_q, j \neq i} \mathbb{1}\{D_{ij}<0\}}{1+\sum_{j\in \Omega_q, j\neq i} \mathbb{1}\{D_{ij}<0\} }.
\end{equation} %
Therefore, with this notation we can directly obtain an smooth approximation making use of Equation~\ref{eq:sigmoid} as
\begin{equation}
    \operatorname{AP}_q \approx \frac{1}{|\mathcal{P}_q|} \sum_{i\in \mathcal{P}_q} \frac{1 + \sum_{j\in \mathcal{P}_q, j \neq i} \mathcal{G}(D_{ij};\tau)}{1+\sum_{j\in \Omega_q, j \neq i} \mathcal{G}(D_{ij};\tau) }.
\end{equation}%
Averaging this approximation for all the queries in the  batch, we can define our loss as %
\begin{equation}
    \mathcal{L}_{AP} = 1 - \frac{1}{Q}\sum_{i=1}^Q AP_q,
    \label{eq:loss_ap}
\end{equation} %
where \(Q\) is the number of queries.

\mysubtitle{Smooth-nDCG}: Following the same idea as above,  we replace the $n$-th position in Equation \ref{eq:dcg} by the ranking function, since it defines the position that the $i$-th element of the retrieved set, and the DCG metric is expressed as %
\begin{equation}
    \operatorname{DCG}_q = \sum_{i\in \Omega_q} \frac{r(i)}{\log_2\left(2 + \sum_{j\in \Omega_q, j \neq i} \mathbb{1}\{D_{ij}<0\}\right)},
\end{equation} %
where $r$ is the same graded function used in Equation~\ref{eq:dcg} but evaluated at element $i$. Therefore, the corresponding smooth approximation is %
\begin{equation}
    \operatorname{DCG}_q \approx \sum_{i\in \Omega_q} \frac{r(i)}{\log_2\left(2 + \sum_{j\in \Omega_q, j \neq i} \mathcal{G}(D_{ij};\tau)\right) }
\end{equation} %
when replacing the indicator function by the sigmoid one.

The smooth-nDCG is then defined by replacing the original $\operatorname{DCG}_q$ by its smooth approximation in Equation~\ref{eq:ndcg} and the loss $\mathcal{L}_{nDCG}$ is defined as %
%
%
\begin{equation}
    \mathcal{L}_{nDCG} = 1 - \frac{1}{Q}\sum_{i=1}^Q nDCG_q,
    \label{eq:loss_dcg}
\end{equation}
where \(Q\) is the number of queries.

\subsection{End-to-End training}

Considering a batch of size \(N_B\) of word images \(X = \{x_i\}_{i=1}^{N_B}\) and their corresponding transcriptions \(Y = \{y_i\}_{i=1}^{N_B}\), the proposed word image and string encoder is trained considering all the elements in both, the query and the retrieval set. Bearing in mind that Smooth-AP cannot be properly used to train the string encoder alone, we combine the Smooth-AP and Smooth-nDCG  loss functions into the following three loss functions: 

\begin{align}
    \mathcal{L}_{\mathrm{img}} &= \mathcal{L}_{AP}(X) + \mathcal{L}_{nDCG}(X) \\
    \mathcal{L}_{\mathrm{str}} &= \mathcal{L}_{nDCG}(Y) \\
    \mathcal{L}_{\mathrm{cross}} &= \mathcal{L}_{AP}(Y, X) + \mathcal{L}_{nDCG}(Y, X)
\end{align}

Moreover, the \(L_1\)-loss \(\mathcal{L}_{L_1}\) between each image embedding and its corresponding word embedding is used to force both models to lay in the same embedding space. This loss is multiplied by a parameter \(\alpha\) that we set experimentally to \(0.5\). Note that the gradients of this loss will only update the weights of the word image encoder. Algorithm~\ref{alg:train} depicts the explained training algorithm in the situation in which both losses are considered. \(\Gamma(\cdot)\) denotes the optimizer function.

\begin{algorithm}[t!]
\hspace*{\algorithmicindent} \textbf{Input:} Input data $\{\mathcal{X}, \mathcal{Y}\}$; alphabet $\mathcal{A}$; max training iterations $T$ \\ 
\hspace*{\algorithmicindent} \textbf{Output: } Networks parameters $\Theta = \{\Theta_{\phi}, \Theta_{\psi}\}$.
\begin{algorithmic}[1]
    \Repeat
    \State Get word images $X=\{x_i\}_{i=1}^{N_B}$ and its corresponding transcription $Y=\{y_i\}_{i=1}^{N_B}$
    \State $\mathcal{L} \leftarrow  \mathcal{L}_{img} + \mathcal{L}_{str} + \mathcal{L}_{cross} + \alpha \frac{1}{N_B} \sum_{i=1}^{N_B} \mathcal{L}_{L_1}$
    \State $\Theta \leftarrow \Theta - \Gamma(\nabla_{\Theta} \mathcal{L})$ \label{alg:line:H}
    \Until{Max training iterations $T$}
\end{algorithmic}
\caption{Training algorithm for the proposed model.} \label{alg:train}
\end{algorithm}

\section{Experimental Evaluation}
\label{s:experiments}

In this section, we present an exhaustive evaluation of the proposed model in the problem of word spotting. All the code necessary to reproduce the experiments is available at \href{http://github.com/priba/ndcg_wordspotting.pytorch}{github.com/priba/ndcg\_wordspotting.pytorch} using the PyTorch framework.

\subsection{Implementation details}

The proposed model is trained for 50 epochs where in each epoch 15,000 samples are drawn by means of a random weighted sampler. This setting ensures that the data is balanced among the different classes during the training of the model. In addition, data augmentation is applied in the form of a random affine transformation. The possible transformations have been kept small to ensure that the text is not altered. Thus, we only allow rotations and shear up to 5 degrees and a scale factor in the range of \(0.9\) and \(1.1\).

For all the experiments the Adam optimizer~\cite{kingma2014adam} has been employed. The learning rate, starting from 1e-4 has been decreased by a factor of 0.25 at epochs 25 and 40.

For evaluation purposes, we used the classic ranking metrics introduced in Section~\ref{s:related}, \ie~mAP and nDCG considering the full retrieval set. In particular, in our setup we define the following relevance function \(r(n)\) to the nDCG metric, these words with edit distances 0, 1, 2, 3 and 4 receive a score of 20, 15, 10, 5 and 3. In addition, the smooth-nDCG loss is trained with the following relevance function, %
\begin{equation}
r(n; \gamma) = \max(0, \gamma - \operatorname{Lev}(q, y_n)),    
\end{equation}%
where \(q\) and \(y_n \in \mathcal{Y}\) are the transcriptions of the query at the \(n\)-th ranking of the retrieval list and \(\operatorname{Lev}(\cdot, \cdot)\) corresponds to the Levenshtein distance~\cite{levenshtein1966binary} between two strings. Moreover, \(\gamma\) is an hyperparameter that has been set experimentally to 4 in our case of study.

\subsection{Experimental discussion}

The proposed model has been evaluated in two different datasets. First, the GW dataset, which is composed of handwriting word images and, second, the IIIT5K dataset, which is composed of words cropped from real scenes.

\mysubtitle{George Washington (GW)~\cite{rath2007word}}. This database is based on handwritten letters written in English by George Washington and his associates during the American Revolutionary War in 1755\footnote{George Washington Papers at the Library of Congress from 1741-1799, Series 2, Letterbook 1, pages 270-279 and 300-309, \url{https://www.loc.gov/collections/george-washington-papers/about-this-collection/}}. It consists of \(20\) pages with  a  total  of \(4,894\) handwritten words. Even though several writers were involved, it presents small variations in style and only minor signs of degradation. There is not official partition for the GW dataset, therefore, we follow the evaluation adopted by Almazan~\etal~\cite{almazan2014word}. Thus, the dataset is splitted in two sets at word level containing $70\%$ of the words for training purpose and the remaining $25\%$ for test. This setting is repeated four times following a cross validation setting. Finally, we report the average result of these four runs.

\mysubtitle{The IIIT 5K-word dataset (IIIT5K) dataset~\cite{mishra2012bmvc}}. This dataset provides 5,000 cropped word images from scene texts and born digital images obtained from Google Image engine search. The authors of this dataset provide an official partition containing two subsets of 2,000 and 3,000 images for training and testing purposes. In addition, the each word is associated with two lexicon subsets of 50 and 1,000 words, respectively. Moreover, a global lexicon~\cite{weinman2009lexicon} of more than half a million words can be used for word recognition. In this work, none of the provided lexicons have been used.

For experimental purposes, we have evaluated three different combinations of the introduced learning objectives, namely, Smooth-AP, Smooth-nDCG and Join,~\ie~a combination of the previous losses.

\begin{table}
    \centering
    \caption{Mean of ranking metrics, mAP and nDCG, for state-of-the-art query-by-string (QbS) and query-by-example (QbE) methods in the GW and IIIT5K datasets.}
    \begin{tabularx}{\textwidth}{X c@{\hskip 4pt}c c@{\hskip 5pt}c c@{\hskip 4pt}c c@{\hskip 6pt}c c@{\hskip 4pt}c c@{\hskip 5pt}c c@{\hskip 4pt}c }
        \toprule
        \textbf{Method} & \multicolumn{6}{c}{\textbf{GW}} &&& \multicolumn{6}{c}{\textbf{IIIT5K}} \\
        \cmidrule{2-7}\cmidrule{10-15}
        & \multicolumn{2}{c}{\textbf{QbS}} &&&
        \multicolumn{2}{c}{\textbf{QbE}} &&& \multicolumn{2}{c}{\textbf{QbS}} &&&
        \multicolumn{2}{c}{\textbf{QbE}} \\
        \cmidrule{2-3}\cmidrule{6-7}\cmidrule{10-11}\cmidrule{14-15}
         & mAP & nDCG &&& mAP & nDCG &&& mAP & nDCG &&& mAP & nDCG \\
        \midrule
        Aldavert~\etal~\cite{aldavert2013integrating} & 56.54 & - &&& - & - &&& - & - &&& - & -\\
        Frinken~\etal~\cite{frinken2011novel} & 84.00 & - &&& - & - &&& - & - &&& - & -\\
        Almaz{\'a}n~\etal~\cite{almazan2014word} & 91.29 & - &&& 93.04 & - &&& 66.24 & - &&& 63.42 & -\\
        Gom{\'e}z~\etal~\cite{gomez2017lsde} & 91.31 & - &&& - & - &&& - & - &&& - & -\\
        Sudholt~\etal~\cite{sudholt2016phocnet} & 92.64 & - &&& 96.71 & - &&& - & - &&& - & -\\
        Krishnan~\etal~\cite{krishnan2016deep} & 92.84 & - &&& 94.41 & - &&& - & - &&& - & -\\
        Wilkinson~\etal~\cite{wilkinson2016semantic} & 93.69 & - &&& \textbf{97.98} & - &&& - & - &&& - & -\\
        Sudholt~\etal~\cite{sudholt2017evaluating} & 98.02 & - &&& 97.78 & - &&& - & - &&& - & -\\
        Krishnan~\etal~\cite{krishnan2018word} & \textbf{98.86} & - &&& 98.01 & - &&& - & - &&& - & - \\
        \midrule
        Smooth-AP~\etal~\cite{brown2020eccv} & 96.79 & 88.99 &&& 97.17 & 87.64 &&& 81.25 & 80.86 &&& 80.60 & 76.13\\
        Smooth-nDCG & 98.25 & \textbf{96.41} &&& 97.94 & \textbf{94.27} &&& 88.99 & 90.63 &&& 87.53 & 85.11\\
        Join & 98.38 & 96.40 &&& \textbf{98.09} & \textbf{94.27} &&& \textbf{89.14} & \textbf{90.64} &&& \textbf{87.60} & \textbf{85.16} \\
        \bottomrule
    \end{tabularx}
    \label{t:results}
\end{table}

Table~\ref{t:results} provides a comparison with the state-of-the-art techniques for word spotting. Most of these techniques have been only evaluated in the handwritten case corresponding to the GW dataset. The best performing model for QbS in the GW dataset is the method proposed by Krishnan~\etal~\cite{krishnan2018word} that obtains a mAP slightly better than our Join setting. Even though, our proposed model has been pre-trained with ImageNet, Krishnan~\etal~makes use of a huge dataset of 9M synthetic word images. Besides the good performance for QbS task, our model is able to perform slightly better in the QbE task. From the same table, we can observe that our model is able to generalize to real scene word images. In such dataset, we outperform the work of Almazán~\etal~by more than 20 points for both tasks, QbS and QbE.

Observe that exploiting the Smooth-nDCG loss, our model is able to enhance the performance of the architecture trained with the Smooth-AP objective function alone. This remarks the importance of not only considering those images whose transcription matches with the query word. In addition, in terms of the nDCG, the performance is also boosted by this loss. Thus, the results are more appealing to the final user following the pre-defined relevance function, in this example, the string edit distance.

\begin{figure}
    \centering
    \setlength\tabcolsep{0pt} 
    \begin{tabular}{ccc}
        \includegraphics[width=0.33\textwidth, trim=20 8 38 40, clip]{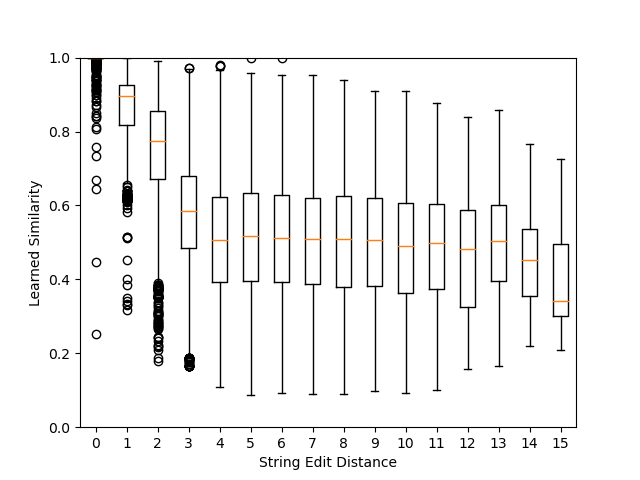} & \includegraphics[width=0.33\textwidth, trim=20 8 38 40, clip]{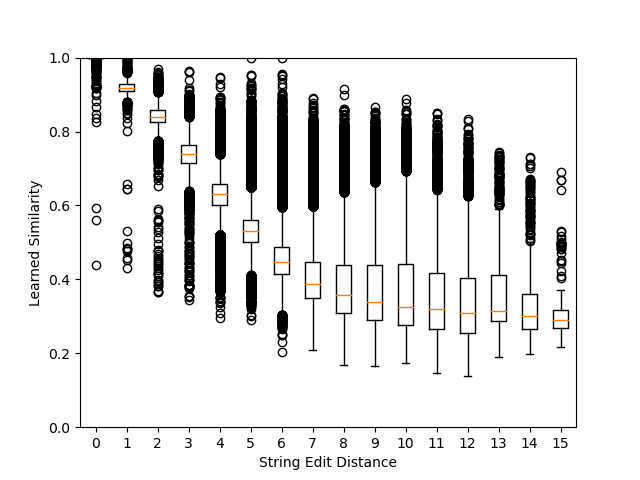} & \includegraphics[width=0.33\textwidth, trim=20 8 38 40, clip]{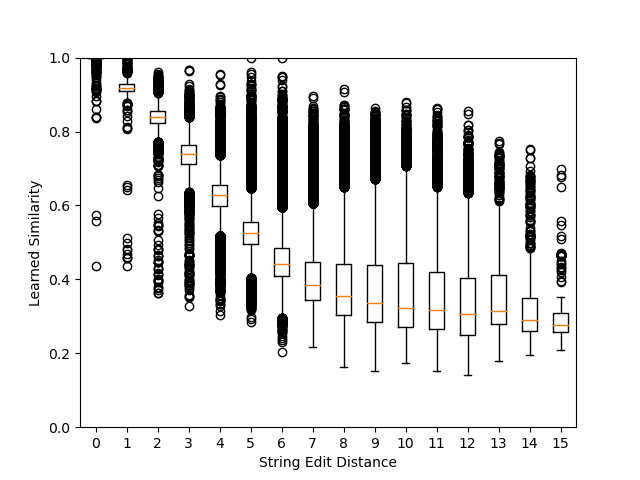} \\
        (a) & (b) & (c)
    \end{tabular}
    \caption{Box plots for (a) smooth-AP; (b) smooth-nDCG; (c) Join, losses.}
    \label{f:boxplot}
\end{figure}

The achieved performance can be explained by the box plots depicted in Figure~\ref{f:boxplot}. This figure shows the correlation between the real string edit distance and the learned similarity of two words. On the one hand, note that the model trained by the smooth-AP loss is able to specialize on detecting which is the matching image given a query. On the other hand, both nDCG or Join losses are able to describe a better ranking when increasing the real string edit distance. 

\begin{figure}[t]
    \centering
    \tiny
    \begin{tabularx}{\textwidth}{>{\centering\arraybackslash}X>{\centering\arraybackslash}X>{\centering\arraybackslash}X>{\centering\arraybackslash}X>{\centering\arraybackslash}X>{\centering\arraybackslash}X>{\centering\arraybackslash}X>{\centering\arraybackslash}X}
    \toprule
        \multicolumn{7}{l}{\footnotesize \textbf{Query:} jones} \\
        \midrule
        \includegraphics[height=0.5cm, width=1.5cm, cfbox=green 1pt 0pt]{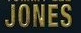} & 
        \includegraphics[height=0.5cm, width=1.5cm, cfbox=green 1pt 0pt]{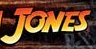} &
        \includegraphics[height=0.5cm, width=1.5cm]{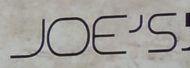} &
        \includegraphics[height=0.5cm, width=1.5cm, cfbox=green 1pt 0pt]{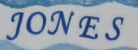} &
        \includegraphics[height=0.5cm, width=1.5cm]{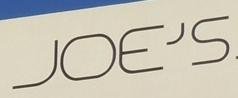} &
        \includegraphics[height=0.5cm, width=1.5cm]{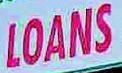} &
        \includegraphics[height=0.5cm, width=1.5cm]{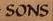} \\
        \(ED=0\) & \(ED=0\) & \(ED=1\) & \(ED=0\) & \(ED=1\) & \(ED=3\) & \(ED=2\) \\
        \midrule
        \multicolumn{7}{l}{\footnotesize \textbf{Query:} bank} \\
        \midrule
        \includegraphics[height=0.5cm, width=1.5cm, cfbox=green 1pt 0pt]{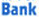} & 
        \includegraphics[height=0.5cm, width=1.5cm, cfbox=green 1pt 0pt]{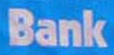} &
        \includegraphics[height=0.5cm, width=1.5cm, cfbox=green 1pt 0pt]{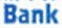} &
        \includegraphics[height=0.5cm, width=1.5cm, cfbox=green 1pt 0pt]{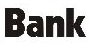} &
        \includegraphics[height=0.5cm, width=1.5cm, cfbox=green 1pt 0pt]{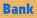} &
        \includegraphics[height=0.5cm, width=1.5cm, cfbox=green 1pt 0pt]{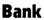} &
        \includegraphics[height=0.5cm, width=1.5cm, cfbox=green 1pt 0pt]{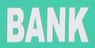} \\
        \(ED=0\) & \(ED=0\) & \(ED=0\) & \(ED=0\) & \(ED=0\) & \(ED=0\) & \(ED=0\) \\
        \midrule
        \multicolumn{7}{l}{\footnotesize \textbf{Query:} advertising} \\
        \midrule
        \includegraphics[height=0.5cm, width=1.5cm, cfbox=green 1pt 0pt]{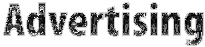} & 
        \includegraphics[height=0.5cm, width=1.5cm, cfbox=green 1pt 0pt]{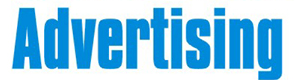} &
        \includegraphics[height=0.5cm, width=1.5cm, cfbox=green 1pt 0pt]{img/qualitative/iiit5k/advertising/2.png} &
        \includegraphics[height=0.5cm, width=1.5cm, cfbox=green 1pt 0pt]{img/qualitative/iiit5k/advertising/3.png} &
        \includegraphics[height=0.5cm, width=1.5cm, cfbox=green 1pt 0pt]{img/qualitative/iiit5k/advertising/4.png} &
        \includegraphics[height=0.5cm, width=1.5cm, cfbox=green 1pt 0pt]{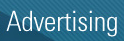} &
        \includegraphics[height=0.5cm, width=1.5cm, cfbox=green 1pt 0pt]{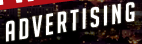} \\
        \(ED=0\) & \(ED=0\) & \(ED=0\) & \(ED=0\) & \(ED=0\) & \(ED=0\) & \(ED=0\) \\
    \bottomrule
    \end{tabularx}
    \caption{Qualitative retrieval results for the IIIT5K dataset using the model trained with the Join loss. In green, the exact matches.}
    \label{f:iiit5k_qualitative}
\end{figure}

\begin{figure}
    \centering
    \tiny
    \begin{tabularx}{\textwidth}{l>{\centering\arraybackslash}X>{\centering\arraybackslash}X>{\centering\arraybackslash}X>{\centering\arraybackslash}X>{\centering\arraybackslash}X>{\centering\arraybackslash}X>{\centering\arraybackslash}X>{\centering\arraybackslash}X}
    \toprule
        \multicolumn{8}{l}{\footnotesize \textbf{Query:} great} \\
        \midrule
        LSDE~\cite{gomez2017lsde} & \includegraphics[height=0.5cm, width=1.5cm, cfbox=green 1pt 0pt]{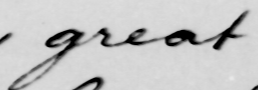} & 
        \includegraphics[height=0.5cm, width=1.5cm, cfbox=green 1pt 0pt]{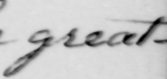} &
        \includegraphics[height=0.5cm, width=1.5cm]{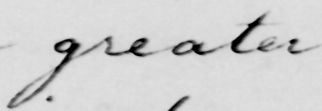} &
        \includegraphics[height=0.5cm, width=1.5cm]{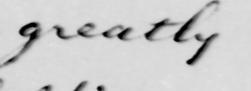} &
        \includegraphics[height=0.5cm, width=1.5cm]{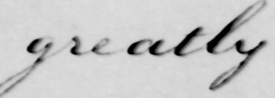} &
        \includegraphics[height=0.5cm, width=1.5cm]{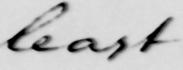} &
        \includegraphics[height=0.5cm, width=1.5cm]{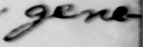} \\
        & \(ED=0\) & \(ED=0\) & \(ED=2\) & \(ED=2\) & \(ED=2\) & \(ED=3\) & \(ED=3\) \\
        \midrule
        S-MAP & \includegraphics[height=0.5cm, width=1.5cm, cfbox=green 1pt 0pt]{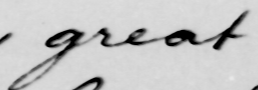} & 
        \includegraphics[height=0.5cm, width=1.5cm, cfbox=green 1pt 0pt]{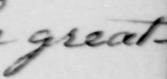} &
        \includegraphics[height=0.5cm, width=1.5cm]{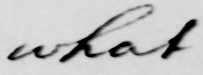} &
        \includegraphics[height=0.5cm, width=1.5cm]{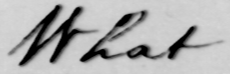} &
        \includegraphics[height=0.5cm, width=1.5cm]{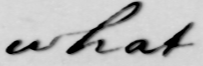} &
        \includegraphics[height=0.5cm, width=1.5cm]{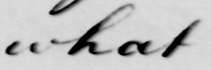} &
        \includegraphics[height=0.5cm, width=1.5cm]{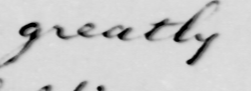} \\
        & \(ED=0\) & \(ED=0\) & \(ED=3\) & \(ED=3\) & \(ED=3\) & \(ED=3\) & \(ED=2\) \\
        \midrule
        S-NDCG & \includegraphics[height=0.5cm, width=1.5cm, cfbox=green 1pt 0pt]{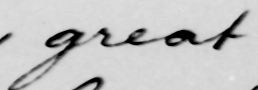} & 
        \includegraphics[height=0.5cm, width=1.5cm, cfbox=green 1pt 0pt]{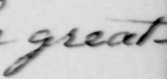} &
        \includegraphics[height=0.5cm, width=1.5cm]{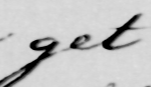} &
        \includegraphics[height=0.5cm, width=1.5cm]{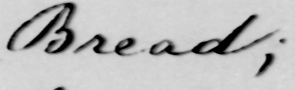} &
        \includegraphics[height=0.5cm, width=1.5cm]{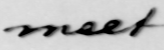} &
        \includegraphics[height=0.5cm, width=1.5cm]{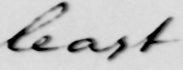} &
        \includegraphics[height=0.5cm, width=1.5cm]{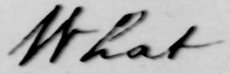} \\
        & \(ED=0\) & \(ED=0\) & \(ED=2\) & \(ED=2\) & \(ED=3\) & \(ED=3\) & \(ED=3\) \\
        \midrule
        Join & \includegraphics[height=0.5cm, width=1.5cm, cfbox=green 1pt 0pt]{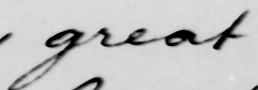} & 
        \includegraphics[height=0.5cm, width=1.5cm, cfbox=green 1pt 0pt]{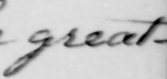} &
        \includegraphics[height=0.5cm, width=1.5cm]{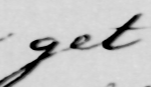} &
        \includegraphics[height=0.5cm, width=1.5cm]{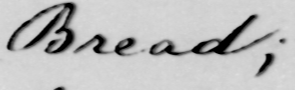} &
        \includegraphics[height=0.5cm, width=1.5cm]{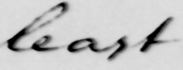} &
        \includegraphics[height=0.5cm, width=1.5cm]{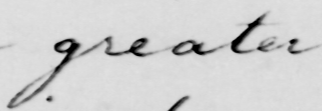} &
        \includegraphics[height=0.5cm, width=1.5cm]{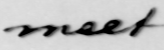} \\
        & \(ED=0\) & \(ED=0\) & \(ED=2\) & \(ED=2\) & \(ED=3\) & \(ED=2\) & \(ED=3\) \\
        \midrule
        \multicolumn{8}{l}{\footnotesize \textbf{Query:} recruits} \\
        \midrule
        LSDE~\cite{gomez2017lsde} & \includegraphics[height=0.5cm, width=1.5cm, cfbox=green 1pt 0pt]{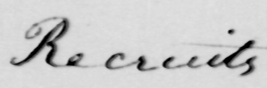} & 
        \includegraphics[height=0.5cm, width=1.5cm, cfbox=green 1pt 0pt]{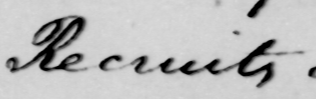} &
        \includegraphics[height=0.5cm, width=1.5cm]{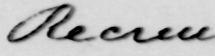} &
        \includegraphics[height=0.5cm, width=1.5cm]{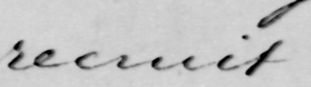} &
        \includegraphics[height=0.5cm, width=1.5cm]{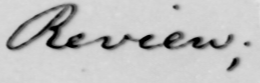} &
        \includegraphics[height=0.5cm, width=1.5cm]{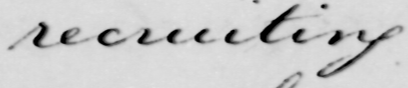} &
        \includegraphics[height=0.5cm, width=1.5cm]{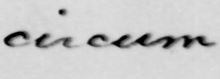} \\
        & \(ED=0\) & \(ED=0\) & \(ED=3\) & \(ED=1\) & \(ED=5\) & \(ED=3\)  & \(ED=7\) \\
        \midrule
        S-MAP & \includegraphics[height=0.5cm, width=1.5cm, cfbox=green 1pt 0pt]{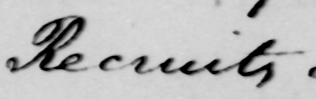} & 
        \includegraphics[height=0.5cm, width=1.5cm, cfbox=green 1pt 0pt]{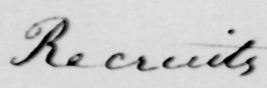} &
        \includegraphics[height=0.5cm, width=1.5cm]{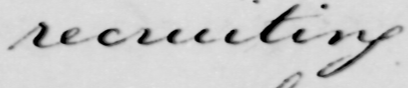} &
        \includegraphics[height=0.5cm, width=1.5cm]{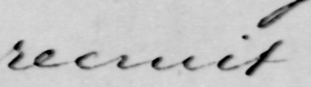} &
        \includegraphics[height=0.5cm, width=1.5cm]{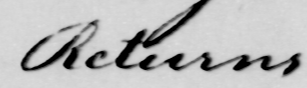} &
        \includegraphics[height=0.5cm, width=1.5cm]{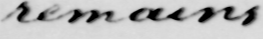} &
        \includegraphics[height=0.5cm, width=1.5cm]{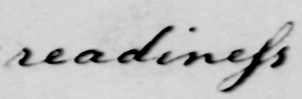} \\
        & \(ED=0\) & \(ED=0\) & \(ED=3\) & \(ED=1\) & \(ED=4\) & \(ED=4\)  & \(ED=6\) \\
        \midrule
        S-NDCG & \includegraphics[height=0.5cm, width=1.5cm, cfbox=green 1pt 0pt]{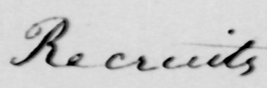} & 
        \includegraphics[height=0.5cm, width=1.5cm, cfbox=green 1pt 0pt]{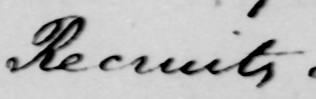} &
        \includegraphics[height=0.5cm, width=1.5cm]{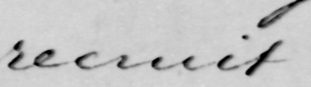} &
        \includegraphics[height=0.5cm, width=1.5cm]{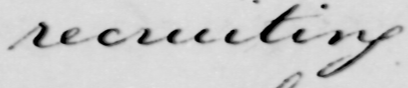} &
        \includegraphics[height=0.5cm, width=1.5cm]{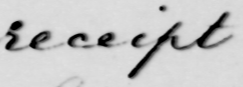} &
        \includegraphics[height=0.5cm, width=1.5cm]{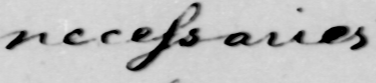} &
        \includegraphics[height=0.5cm, width=1.5cm]{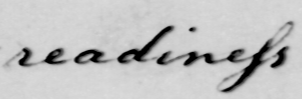} \\
        & \(ED=0\) & \(ED=0\) & \(ED=1\) & \(ED=3\) &  \(ED=4\) & \(ED=7\)  & \(ED=6\) \\
        \midrule
        Join & \includegraphics[height=0.5cm, width=1.5cm, cfbox=green 1pt 0pt]{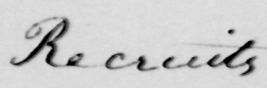} & 
        \includegraphics[height=0.5cm, width=1.5cm, cfbox=green 1pt 0pt]{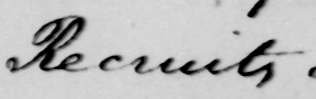} &
        \includegraphics[height=0.5cm, width=1.5cm]{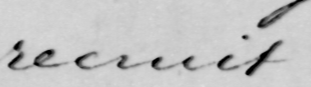} &
        \includegraphics[height=0.5cm, width=1.5cm]{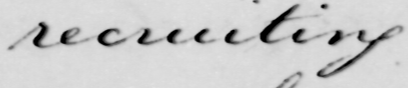} &
        \includegraphics[height=0.5cm, width=1.5cm]{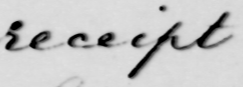} &
        \includegraphics[height=0.5cm, width=1.5cm]{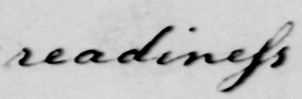} &
        \includegraphics[height=0.5cm, width=1.5cm]{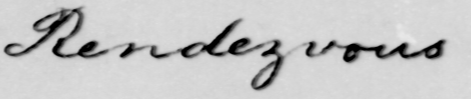} \\
        & \(ED=0\) & \(ED=0\) & \(ED=1\) & \(ED=3\) &  \(ED=4\) & \(ED=6\) & \(ED=7\) \\
        \midrule
        \multicolumn{8}{l}{\footnotesize \textbf{Query:} honour} \\
        \midrule
        LSDE~\cite{gomez2017lsde} & \includegraphics[height=0.5cm, width=1.5cm, cfbox=green 1pt 0pt]{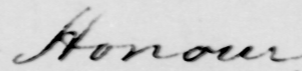} & 
        \includegraphics[height=0.5cm, width=1.5cm, cfbox=green 1pt 0pt]{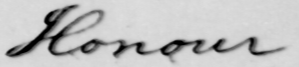} &
        \includegraphics[height=0.5cm, width=1.5cm]{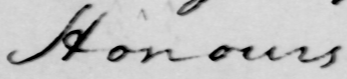} &
        \includegraphics[height=0.5cm, width=1.5cm]{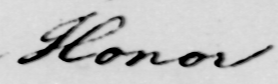} &
        \includegraphics[height=0.5cm, width=1.5cm, cfbox=green 1pt 0pt]{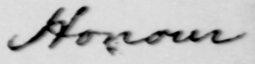} &
        \includegraphics[height=0.5cm, width=1.5cm]{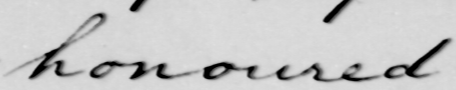} &
        \includegraphics[height=0.5cm, width=1.5cm]{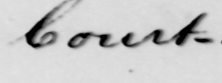} \\
        & \(ED=0\) & \(ED=0\) & \(ED=1\) & \(ED=1\) &  \(ED=0\) & \(ED=2\) & \(ED=4\) \\
        \midrule
        S-MAP & \includegraphics[height=0.5cm, width=1.5cm, cfbox=green 1pt 0pt]{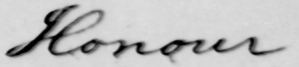} & 
        \includegraphics[height=0.5cm, width=1.5cm, cfbox=green 1pt 0pt]{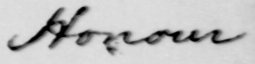} &
        \includegraphics[height=0.5cm, width=1.5cm, cfbox=green 1pt 0pt]{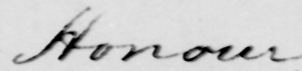} &
        \includegraphics[height=0.5cm, width=1.5cm]{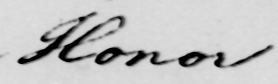} &
        \includegraphics[height=0.5cm, width=1.5cm]{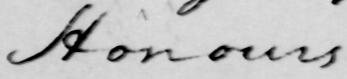} &
        \includegraphics[height=0.5cm, width=1.5cm]{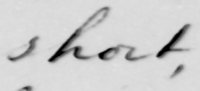} &
        \includegraphics[height=0.5cm, width=1.5cm]{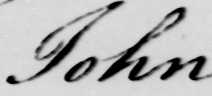} \\
        & \(ED=0\) & \(ED=0\) & \(ED=0\) & \(ED=1\) &  \(ED=1\) & \(ED=5\) & \(ED=5\) \\
        \midrule
        S-NDCG & \includegraphics[height=0.5cm, width=1.5cm, cfbox=green 1pt 0pt, cfbox=green 1pt 0pt]{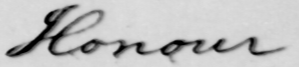} & 
        \includegraphics[height=0.5cm, width=1.5cm, cfbox=green 1pt 0pt]{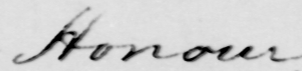} &
        \includegraphics[height=0.5cm, width=1.5cm, cfbox=green 1pt 0pt]{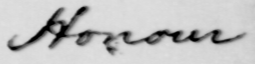} &
        \includegraphics[height=0.5cm, width=1.5cm]{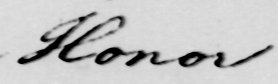} &
        \includegraphics[height=0.5cm, width=1.5cm]{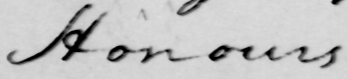} &
        \includegraphics[height=0.5cm, width=1.5cm]{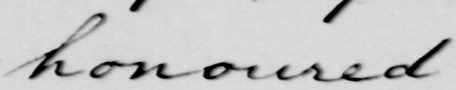} &
        \includegraphics[height=0.5cm, width=1.5cm]{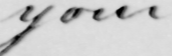} \\
        & \(ED=0\) & \(ED=0\) & \(ED=0\) & \(ED=1\) &  \(ED=1\) & \(ED=2\) & \(ED=3\) \\
        \midrule
        Join & \includegraphics[height=0.5cm, width=1.5cm, cfbox=green 1pt 0pt]{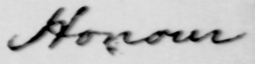} & 
        \includegraphics[height=0.5cm, width=1.5cm, cfbox=green 1pt 0pt]{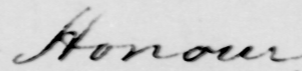} &
        \includegraphics[height=0.5cm, width=1.5cm, cfbox=green 1pt 0pt]{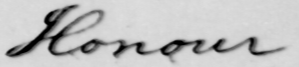} &
        \includegraphics[height=0.5cm, width=1.5cm]{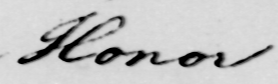} &
        \includegraphics[height=0.5cm, width=1.5cm]{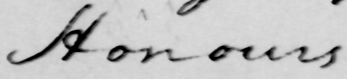} &
        \includegraphics[height=0.5cm, width=1.5cm]{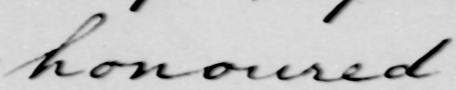} &
        \includegraphics[height=0.5cm, width=1.5cm]{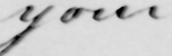} \\
        & \(ED=0\) & \(ED=0\) & \(ED=0\) & \(ED=1\) &  \(ED=1\) & \(ED=2\) & \(ED=3\) \\
        \midrule
        \multicolumn{8}{l}{\footnotesize \textbf{Query:} deliver} \\
        \midrule
        LSDE~\cite{gomez2017lsde} & \includegraphics[height=0.5cm, width=1.5cm, cfbox=green 1pt 0pt]{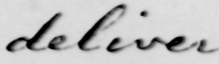} & 
        \includegraphics[height=0.5cm, width=1.5cm, cfbox=green 1pt 0pt]{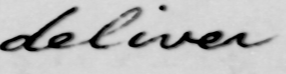} &
        \includegraphics[height=0.5cm, width=1.5cm]{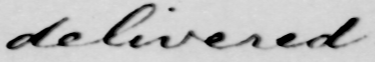} &
        \includegraphics[height=0.5cm, width=1.5cm]{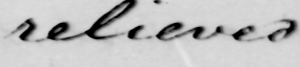} &
        \includegraphics[height=0.5cm, width=1.5cm]{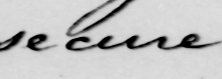} &
        \includegraphics[height=0.5cm, width=1.5cm]{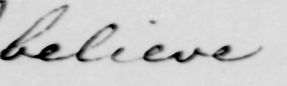} &
        \includegraphics[height=0.5cm, width=1.5cm]{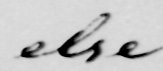} \\
        & \(ED=0\) & \(ED=0\) & \(ED=2\) & \(ED=3\) &  \(ED=5\) & \(ED=3\) & \(ED=4\) \\
        \midrule
        S-MAP & \includegraphics[height=0.5cm, width=1.5cm, cfbox=green 1pt 0pt]{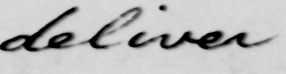} & 
        \includegraphics[height=0.5cm, width=1.5cm, cfbox=green 1pt 0pt]{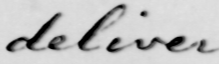} &
        \includegraphics[height=0.5cm, width=1.5cm]{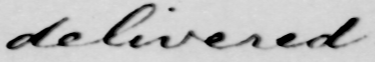} &
        \includegraphics[height=0.5cm, width=1.5cm]{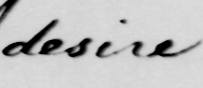} &
        \includegraphics[height=0.5cm, width=1.5cm]{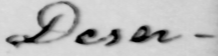} &
        \includegraphics[height=0.5cm, width=1.5cm]{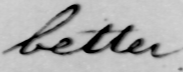} &
        \includegraphics[height=0.5cm, width=1.5cm]{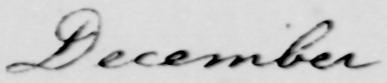} \\
        & \(ED=0\) & \(ED=0\) & \(ED=2\) & \(ED=3\) &  \(ED=3\) & \(ED=4\) & \(ED=4\) \\
        \midrule
        S-NDCG & \includegraphics[height=0.5cm, width=1.5cm, cfbox=green 1pt 0pt]{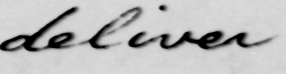} & 
        \includegraphics[height=0.5cm, width=1.5cm, cfbox=green 1pt 0pt]{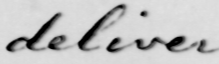} &
        \includegraphics[height=0.5cm, width=1.5cm]{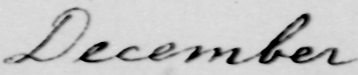} &
        \includegraphics[height=0.5cm, width=1.5cm]{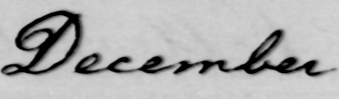} &
        \includegraphics[height=0.5cm, width=1.5cm]{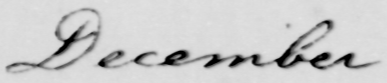} &
        \includegraphics[height=0.5cm, width=1.5cm]{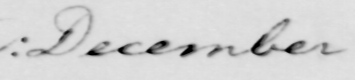} &
        \includegraphics[height=0.5cm, width=1.5cm]{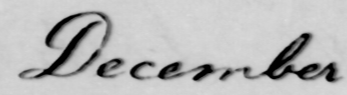} \\
        & \(ED=0\) & \(ED=0\) & \(ED=4\) & \(ED=4\) & \(ED=4\) &  \(ED=4\) & \(ED=4\) \\
        \midrule
        Join & \includegraphics[height=0.5cm, width=1.5cm, cfbox=green 1pt 0pt]{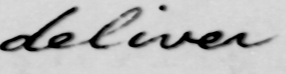} & 
        \includegraphics[height=0.5cm, width=1.5cm, cfbox=green 1pt 0pt]{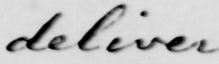} &
        \includegraphics[height=0.5cm, width=1.5cm]{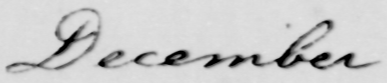} &
        \includegraphics[height=0.5cm, width=1.5cm]{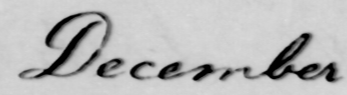} &
        \includegraphics[height=0.5cm, width=1.5cm]{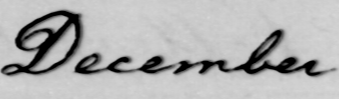} &
        \includegraphics[height=0.5cm, width=1.5cm]{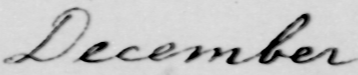} &
        \includegraphics[height=0.5cm, width=1.5cm]{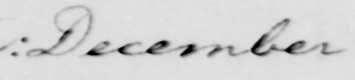} \\
        & \(ED=0\) & \(ED=0\) & \(ED=4\) & \(ED=4\) & \(ED=4\) &  \(ED=4\) & \(ED=4\) \\
    \bottomrule
    \end{tabularx}
    \caption{Qualitative results for the GW dataset. In green, the exact matches.}
    \label{f:gw_qualitative}
\end{figure}

Figures~\ref{f:iiit5k_qualitative} and~\ref{f:gw_qualitative} demonstrates qualitatively the performance of the proposed setting. Observe that the proposed model is able to rank in the first positions the exact match given the query word. Moreover, introducing the Smooth-nDCG we observe that the edit distance is in general smaller than in the Smooth-AP case.

\begin{figure}
    \centering
    \includegraphics[width=0.6\textwidth]{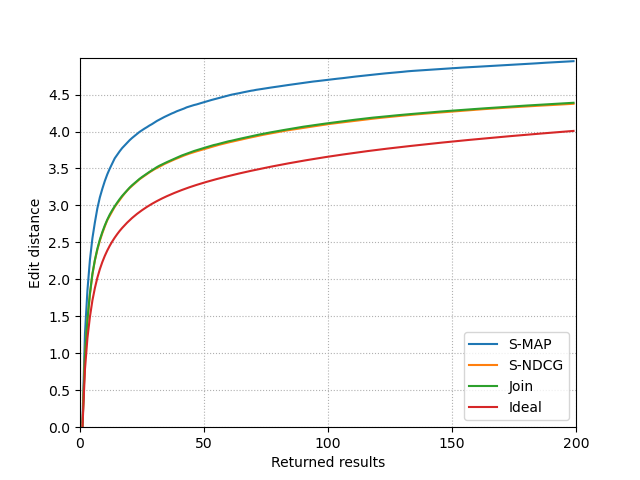}
    \caption{Average edit distance among the top-\(n\) returned results. Note that Smooth-nDCG and Join overlaps in the plot being Smooth-nDCG slightly closer to the Ideal case.}
    \label{f:distance}
\end{figure}

Finally, Figure~\ref{f:distance} provides an overview of the average edit distance of the top-\(n\) results given a query. For instance, the Ideal case, tells us that considering the ground-truth, in average, all the retrieved images in the top-50 have an edit distance smaller than 3.5. Here, we can clearly identify that both learning objectives Smooth-nDCG and Join are able to close the gap between the Ideal case and the Smooth-AP loss.

\section{Conclusions}
\label{s:conclusions}

In this work, we have presented a word spotting framework completely based on ranking-based losses. The proposed approach learns directly from the retrieval list rather than pairs or triplets as most of the state-of-the-art methodologies. In addition, we do not require any prelearned word embedding. From the application point of view, we have shown the competitive performance of our model against the state-of-the-art methods for word spotting in handwritten and real scene text images. Overall, we have demonstrated the importance of considering not only the corresponding image/transcription pair but also, they relation between the different elements in the batch thanks to a graded relevance score.

As future work, we plan to perform an exhaustive evaluation on the different hyperparameters of the proposed smooth-nDCG objective such as, the relevance and the indicator functions. Moreover, the final loss can be weighted between the smooth-AP and the smooth-nDCG losses. Finally, we would like to explore how this framework extends to other multi-modal retrieval tasks.

\section*{Acknowledgment}

This work has been partially supported by the Spanish projects RTI2018-095645-B-C21, and FCT-19-15244, and the Catalan projects 2017-SGR-1783, the Culture Department of the Generalitat de Catalunya, and the CERCA Program / Generalitat de Catalunya.

\bibliographystyle{splncs04}
\bibliography{bib/bibliography.bib}

\end{document}